%% file: main.tex
\documentclass[10pt,twocolumn,letterpaper]{article}

\usepackage[pagenumbers]{cvpr} 
\usepackage{gensymb}
\input{preamble}

\usepackage{graphicx}
\usepackage{amsmath}
\usepackage{amssymb}
\usepackage{booktabs}

\definecolor{cvprblue}{rgb}{0.21,0.49,0.74}
\usepackage[pagebackref,breaklinks,colorlinks,citecolor=cvprblue]{hyperref}

\usepackage[capitalize]{cleveref}
\crefname{section}{Sec.}{Secs.}
\Crefname{section}{Section}{Sections}
\Crefname{table}{Table}{Tables}
\crefname{table}{Tab.}{Tabs.}

\def\paperID{3306} 
\def\confName{CVPR}
\def\confYear{2024}

\include{our_cmds}

\begin{document}

\title{Slice3D: Multi-Slice, Occlusion-Revealing, Single View 3D Reconstruction}

\author{Yizhi Wang$^1$, Wallace Lira$^1$, Wenqi Wang$^2$, Ali Mahdavi-Amiri$^1$, Hao Zhang$^1$\\
$^1$Simon Fraser University
$^2$Tsinghua University}

\twocolumn[{
\maketitle
\vspace{-2em}
\centering
\includegraphics[width=.95\textwidth]{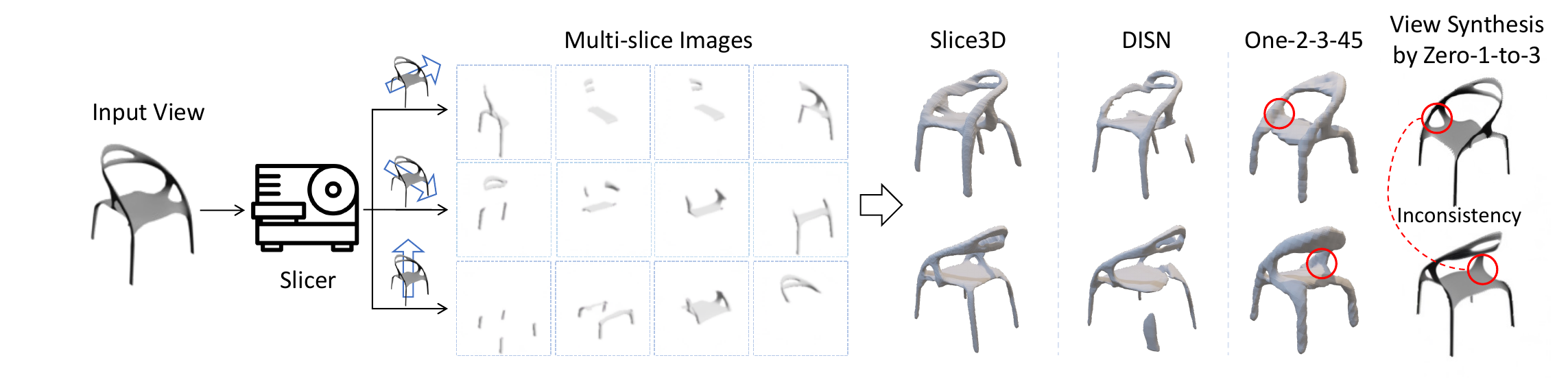}
\captionof{figure}{
Our single-view 3D reconstruction method, Slice3D, predicts {\em multi-slice images\/} to reveal occluded parts without changing the camera (in contrast to {\em multi-view\/} synthesis), and then lifts the slices into a 3D model. This leads to better recovery of occluded parts than both encoder-decoder based methods such as DISN~\cite{xu2019disn} and 3D reconstructions from diffusion-synthesized multi-view images (via Zero-1-to-3~\cite{liu2023zero}) such as One-2-3-45~\cite{liu2023one}. The key challenge of the latter is multi-view consistency, or lack thereof, as shown on the right. 
}
\vspace{1em}
\label{fig:teaser}
}]

\begin{abstract}
We introduce {\em multi-slice\/} reasoning, a new notion for single-view 3D reconstruction which challenges the current and 
prevailing belief that multi-view synthesis is the most natural conduit between single-view and 3D.
Our key observation is that object slicing is more advantageous 
than altering views to reveal occluded structures.
Specifically, slicing 
can peel through any occluder without obstruction, and in the limit (infinitely many slices), it is guaranteed to unveil all hidden object parts.
We realize our idea by developing Slice3D, a novel method for single-view 3D reconstruction 
which first predicts multi-slice {\em images\/} from a single RGB image and then integrates the slices into a 3D model using a coordinate-based transformer network for signed distance prediction.
The slice images can be regressed or generated, both through a U-Net based network. For the former, we inject a learnable slice indicator code 
to designate each decoded image into a spatial slice location, while the slice generator is a denoising diffusion model operating on the entirety of slice images stacked on the input channels. 
%
%
We conduct extensive evaluation against state-of-the-art alternatives to demonstrate superiority of our method, especially in recovering complex and severely occluded shape structures, amid ambiguities.
All Slice3D results were produced by networks trained on a {\em single\/} Nvidia A40 GPU, with an inference time less than 20 seconds.
\end{abstract}

\input{00_intro}
\input{01_RW}
\input{02_method}
\input{03_results}
\input{04_conclusion}
\input{05_supp}

\clearpage
{\small
\bibliographystyle{ieee_fullname}
\bibliography{egbib}
}

\end{document}

%% file: preamble.tex
\usepackage[dvipsnames]{xcolor}


%% file: our_cmds.tex
\usepackage{color}
\definecolor{green}{rgb}{0, 0.5, 0}
\definecolor{orange}{rgb}{0.8, 0.6, 0.2}
\definecolor{red}{rgb}{1.0, 0.0, 0.0}
\definecolor{teal}{rgb}{0.0, 0.4, 0.4}
\definecolor{purple}{rgb}{0.65,0,0.65}
\definecolor{saffron}{rgb}{0.95,0.75,0.2}
\definecolor{turquoise}{rgb}{0.0,0.5,0.5}
\definecolor{black}{rgb}{0.0, 0.0, 0.0}
\definecolor{gray}{rgb}{0.5, 0.5, 0.5}

\newcommand{\am}[1]{{\color{orange}#1}}

\newcommand{\yw}[1]{{\color{green}#1}}

\newcommand{\mypara}[1]{\noindent \textbf{\textit{#1}.}}

%% file: 00_intro.tex
\section{Introduction}
\label{sec:intro}

Single-view 3D reconstruction has been one of the extensively studied problems in computer vision. Despite the rapid 
advances on learning-based approaches to tackle this problem, including the latest attempts to leverage large 
foundational models and powerful diffusion-based generators, the most fundamental challenge still
remains: how to faithfully reconstruct {\em occluded\/} parts from just one view?

\begin{figure}[b]
    \centering
    \includegraphics[width=0.98\columnwidth]{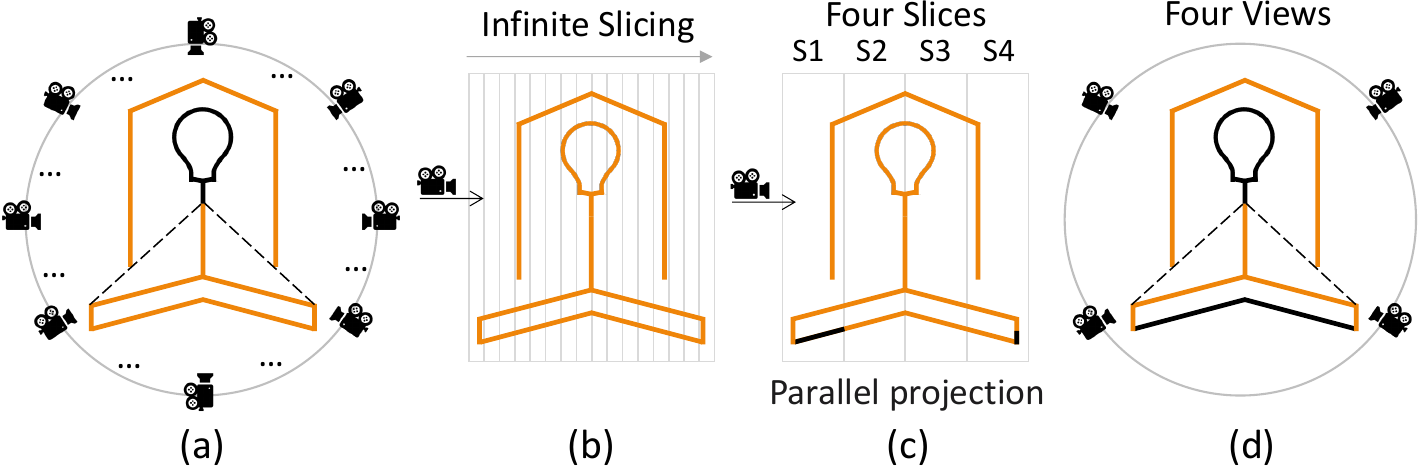}
\caption{Multi-view vs.~multi-slice for occlusion revelation, with \textcolor{orange}{orange} edges showing revealed parts and black edges showing occluded parts. In the limit, multi-view (i.e., infinitely many views)
may still leave undisclosed parts (a), e.g., the bulb, while infinitely many slices would {\em guarantee complete structure revelation\/} (b). With limited slices (c) and views (d), four each, the multi-slice
approach tends to reveal more shape structures.}
\label{fig:lamp_ill}
\end{figure}

\begin{figure*}[!ht]
    \centering
    \includegraphics[width=0.97\textwidth]{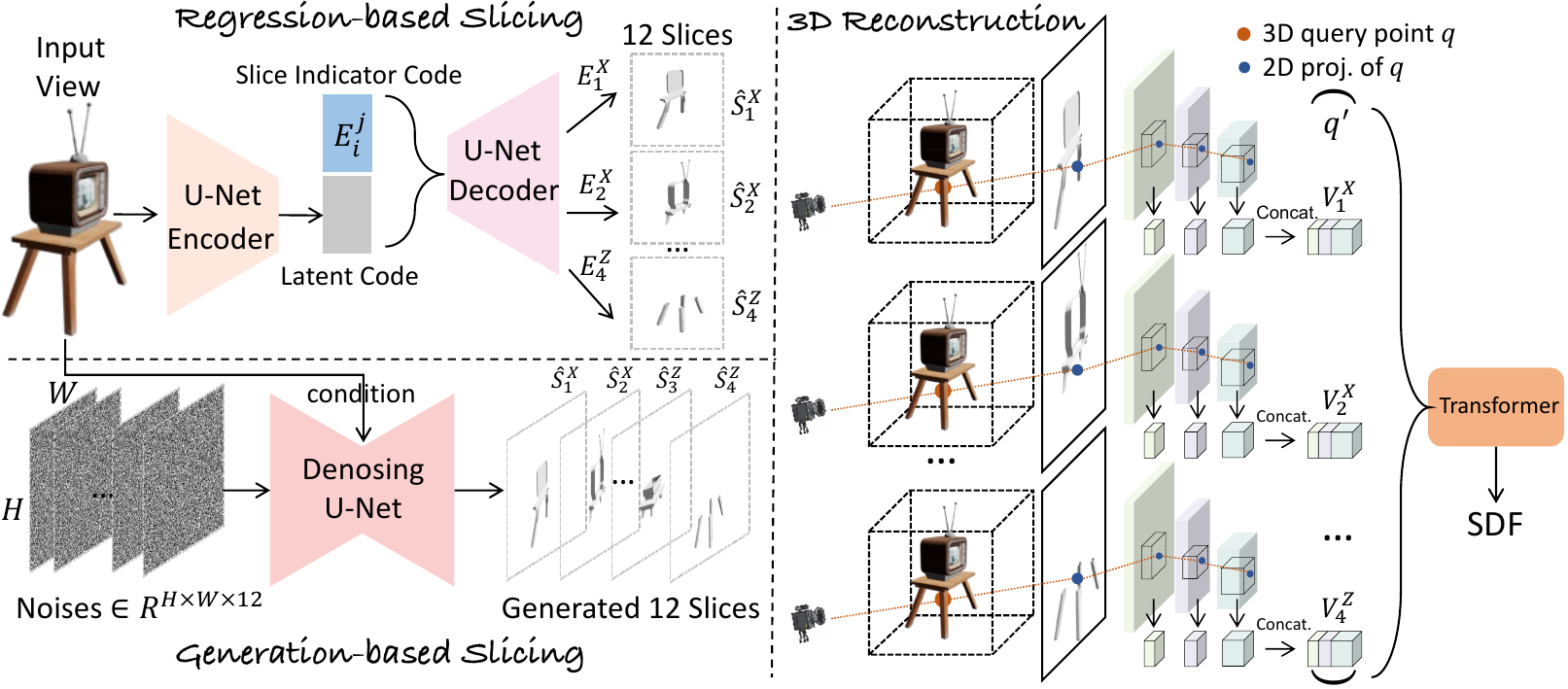}
\caption{Pipeline of Slice3D, for single-view 3D reconstruction from multi-slice images. The illustrated example shows 4 slices per $X$, $Y$, $Z$ direction. The slice images can be obtained by either a regressor or a generator (details in Section~\ref{sec:method_gen_slices}). Regression is carried out by a U-Net encoder-decoder, where the decoder takes an extra {\em slice indicator code\/} that is appended to the encoder output, to produce the corresponding slice image. There is one indicator code for each of the $3 \times 4 = 12$ slices, that is optimized during training. Our generator is a denoising diffusion model that is trained to produce a concatenation of slice images following the distribution of the ground-truth (GT) images, when conditioned on the single-view input. All images are of dimension $H \times W$. Inference from random noises may output multiple plausible results. The 3D reconstruction stage takes as input the predicted slice images to produce an implicit field; see Section~\ref{sec:method_learn_imp_field}. We first extract hierarchical CNN features from the input slices, then project a 3D query point onto the 2D planes of the feature maps, and finally gather all the query-related features and send them into a transformer~\cite{vaswani2017attention} to predict the signed distance at the query point.
}
\label{fig:pipeline}
\end{figure*}

Conventional encoder-decoder based methods, e.g.,~\cite{atlasnet,octreeNet,matryoshka,Chen_2020_CVPR,xu2019disn,li_cvpr21}, 
rely on direct 3D supervision to learn mappings from global or local image features to 3D, but such features 
from just a single view often do not have the capacity to handle severe occlusion.
Also, recent evidence suggests that many of these methods are essentially 
performing image classification or retrieval~\cite{what3d_cvpr19}. As such, occlusion recovery is by 
and large a ``hit or miss''. 
Differentiable rendering enables 3D reasoning from images without 3D supervision~\cite{chen2019dib,dvr2020}, 
but image-space rendering losses alone do not account for occluded structures.

\input{figs/fig_slicer_rep}

Recently, more and more works model the reconstruction problem as a single-view conditioned \textit{generative} task, aiming to learn a distribution 
of unseen structures. Typically, {\em multi-view\/} images are generated from single view via diffusion. This is followed
by multi-view reconstruction, e.g., 
via neural radiance fields (NeRF)~\cite{chen2023single,liu2023one,pavllo2023shape}, to obtain 3D outputs.
However, multi-view consistency is a reoccurring challenge (see Fig.~\ref{fig:teaser}), especially over disparate views.

In this paper, we introduce a novel approach to single-view 3D reconstruction which is
{\em occlusion-revealing\/} (see Fig.~\ref{fig:lamp_ill}) by first predicting {\em multi-slice images\/} from the single-view input and then
integrating the slices into a 3D model, as illustrated in Fig.~\ref{fig:pipeline}. 
In Fig.~\ref{fig:slicer_rep}, we show slice images provided during training 
along a set of slicing directions. 
Each slice represents the projected image, in the input camera 
view, of one slice of the 3D {\em mesh\/} model. For the TV example shown, one can observe that a
completely occluded leg, in the camera view, is 
revealed in the slice images.

The motivations behind our multi-slice reconstruction are three-fold.
First, slicing through a volume or 3D mesh is a natural way to reveal occluded parts.
Second, with the slice images forming an intermediate representation, our approach breaks the difficult problem of mapping from 
single view to 3D into two comparatively simpler tasks: slice prediction followed by 3D reconstruction from slices.

More importantly, when contrasting our multi-{\em slice\/} representation with the popular multi-{\em view\/} approach for occlusion revelation,
we observe that slicing in the limit (i.e., with infinitely many slices) is {\em guaranteed\/} to unveil all hidden structures, while the same 
cannot be said about multi-view captures, as illustrated in Fig.~\ref{fig:lamp_ill} (a-b), for a 2D lamp example. In practice, when slice and view
counts are both limited, the slicing approach is more occlusion-revealing in general since it can peel through any occluders without any obstruction, unlike the 
multi-view case; see (c-d).
On the technical front, neither our slice prediction step nor the 3D reconstruction step is 
concerned with camera view changes or multi-view consistency, which facilitates the use of convolutional features. 
In Fig.~\ref{fig:rec_interiors}, we show the above contrasting characteristics of the multi-slice vs.~multi-view approaches manifested 
in a real 3D lamp reconstruction. 



As outlined in Fig.~\ref{fig:pipeline}, our slice images can be produced by either a regressive network or a generative one, both
employing a U-Net architecture, with network losses measured against ground-truth (GT) images obtained by slicing 3D models from
the training set; see Fig.~\ref{fig:slicer_rep}. In the regressive network, we inject a learnable slice indicator code into the 
encoder-decoder setup to designate each decoded image into the proper spatial slice location. The slice generator is a denoising
diffusion model operating on separate channels for different slice images. In the second stage, 3D reconstruction from the obtained
slice images, we design a coordinate-based transformer to obtain an implicit field.

\input{figs/fig_rec_interiors}

Our network is coined Slice3D. We conduct extensive experiments on ShapeNet~\cite{chang2015shapenet} and Objverse~\cite{deitke2023objaverse}
for evaluation.
Quantitative and qualitative comparisons are made to representative methods including encoder-decoders capturing local image features~\cite{xu2019disn}, autoregressive latent models~\cite{mittal2022autosdf}, 
generative model inversion~\cite{pavllo2023shape}, and three recent methods with diffusion + NeRF~\cite{chen2023single,liu2023one,melas2023realfusion},
demonstrating superiority of our method in terms of reconstruction quality, 
generalizability, and handling of ambiguities.

%% file: figs/fig_slicer_rep.tex
\addtolength{\textfloatsep}{-0.2in}
\begin{figure}[!t]
    \centering
    \includegraphics[width=0.82\columnwidth]{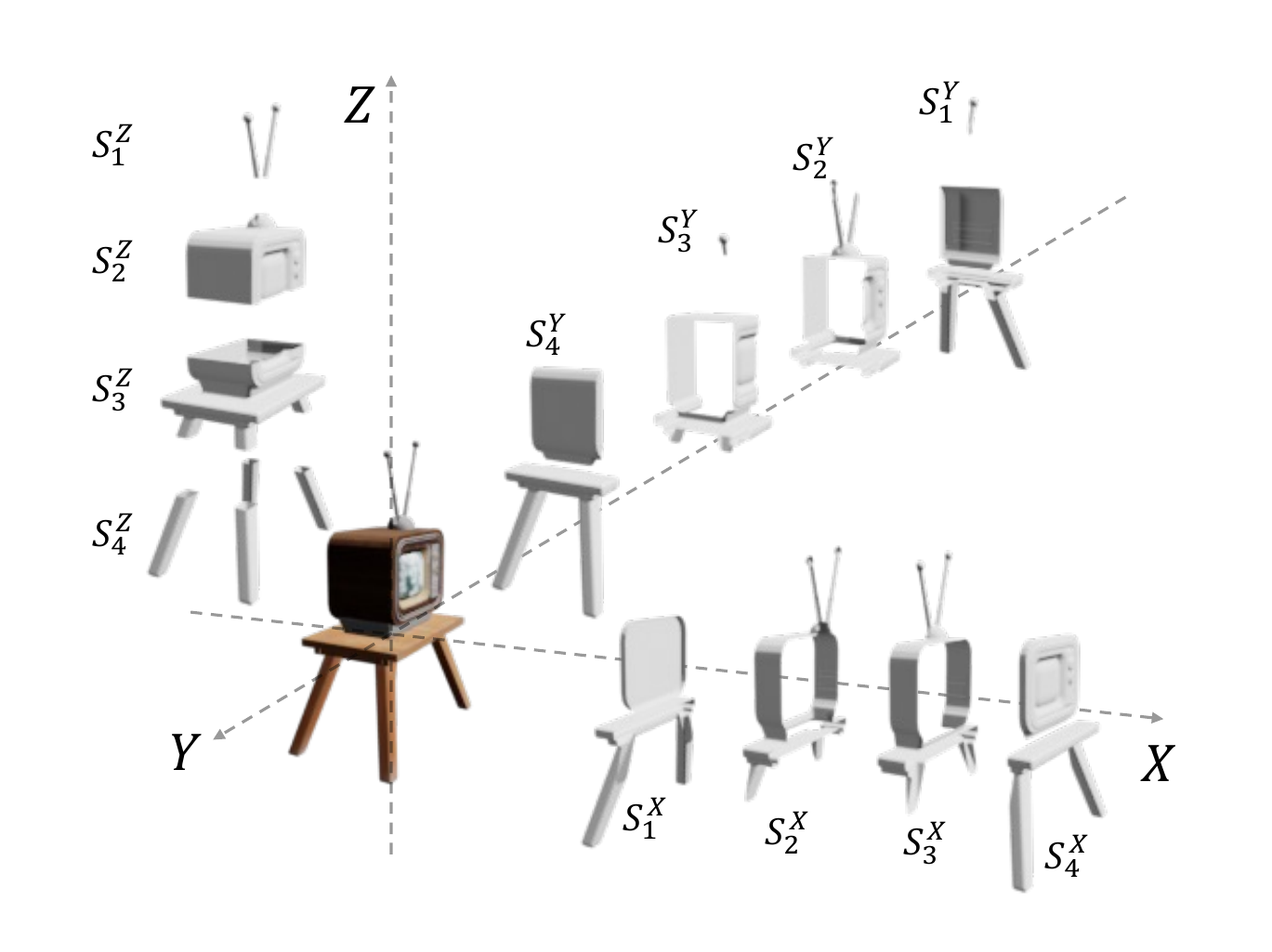}
\caption{Our slice image prediction networks are trained using occlusion-revealing multi-slice images produced from a GT 3D object {\em mesh\/} along a predetermined set of slicing directions (the $X$, $Y$, $Z$ axes of the object bounding box in the example above),  while keeping the hollows as holes in the images, through which the {\em back faces\/} of the mesh can be visible. The 3D mesh may be textured, but the slice images are all rendered without texture.}
\label{fig:slicer_rep}
\end{figure}

%% file: figs/fig_rec_interiors.tex
\begin{figure}[!t]
    \centering
    \includegraphics[width=0.90\columnwidth]{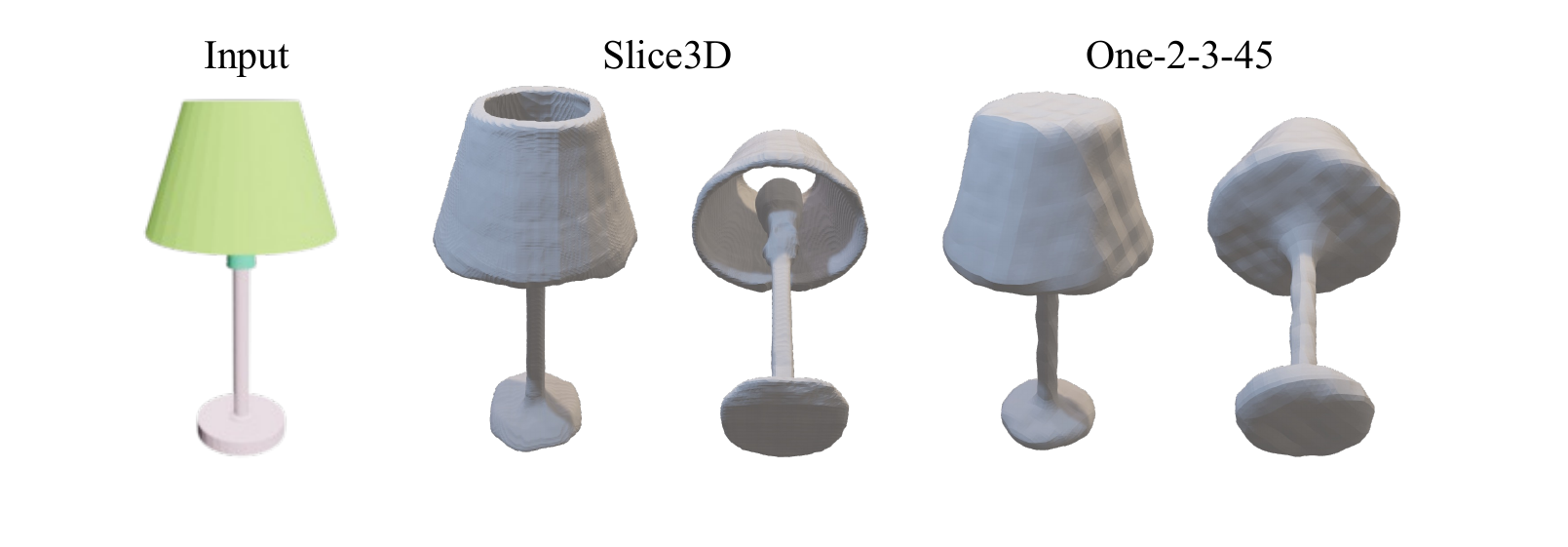}
\caption{Singe-view reconstruction 
of severely occluded parts in a lamp (e.g., the bulb) by Slice3D vs.~One-2-3-45~\cite{liu2023one}, a state-of-the-art multi-view approach. The multi-view images in~\cite{liu2023one} were obtained by Zero-1-to-3~\cite{liu2023zero}, which was trained on
renderings of 800K 3D models~\cite{deitke2023objaverse}. It is conceivable that with a limited number of views (12, the same as our slice count), the bulbs were barely visible due to occlusion by the lamp shades and bases.}
\label{fig:rec_interiors}
\end{figure}

%% file: 01_RW.tex
\section{Related work}
\label{sec:related}


The earliest inspiration for our multi-slice 3D reconstruction idea was GANHopper by Lira et al.~\cite{lira_eccv20}. In the context of unsupervised image-to-image translation, the key idea
of their work is to {\em gradually\/} execute the translation between disparate domains by making small ``hops,'' rather than completing it in one go. We were then motivated to tackle the single-view
3D reconstruction problem as a domain translation task, hopping through the multi-slice images so as to bridge the gap between 2D and 3D.

\vspace{2pt}

\mypara{Neural fields}
Neural implicit functions~\cite{NF_survey} are learned by a neural network to predict such properties as color, occupancy, opacity, and surface normals.
Earlier works such as IM-Net~\cite{chen2019learning}, occupancy networks~\cite{mescheder2019occupancy}, and DeepSDF~\cite{park2019deepsdf} represent a 3D shape as a latent code and send it to 
a decoder formed by MLPs to predict occupancy or SDF values for any query point sampled in 3D.
NeRF~\cite{mildenhall2021nerf} and its variants~\cite{gao2022nerf} represent a 3D scene using a neural radiance field 
for novel view synthesis.
Many works on neural 3D reconstruction~\cite{NF_survey,gao2022nerf} take on single- or multi-view images as well as point clouds as input. In contrast, the reconstruction module in Slice3D is built on
multi-slice images.




\vspace{2pt}

\mypara{Single-view reconstruction from image features}
An early work to exploit neural image features for 3D reconstruction was Pixel2Mesh~\cite{wang2018pixel2mesh}, which uses a graph representation.
Subsequently, neural implicit fields have been widely adopted for single-view reconstruction.
IM-Net~\cite{chen2019learning} 
first extracts a global latent code, from an input view, to learn a 3D implicit field.
Auto-SDF~\cite{mittal2022autosdf} and ShapeFormer~\cite{yan2022shapeformer} employ vector quantization (VQ)~\cite{van2017neural} to improve the representation of latent features.
More recent works~\cite{zhang20233dshape2vecset,nichol2022point,jun2023shap} consider single-view reconstruction as a 3D generation task conditioned on the latent code of the input view.
However, a global latent code is insufficient to capture finer details on a 3D object, often leading to overly smoothed or retrieved outcomes that do not faithfully represent the input.
To alleviate this problem, both DISN~\cite{xu2019disn} and D$^2$IM-Net~\cite{li_cvpr21} opted to learn local images features, along with global ones, to improve the recovery of fine-grained topological (e.g., thin)
features and surface details, respectively.

\vspace{2pt}

\mypara{3D reconstruction from synthesized multi-views}
Conventional 3D reconstruction approaches~\cite{furukawa2015multi} involve leveraging multi-view reasonings. 
With recent advances in generative AI, a prevailing practice in single-view 3D reconstruction is to first employ generative models such as GANs~\cite{goodfellow2014generative,karras2019style} and diffusion models~\cite{sohl2015deep,ho2020denoising,dhariwal2021diffusion,rombach2022high} to generate several novel views from the input view, and then feed them to a multi-view 3D reconstruction module such as NeRF.
NeRF-based surface reconstruction techniques~\cite{wang2021neus,oechsle2021unisurf,yariv2021volume} can be further employed to output 3D mesh models.
EG3D~\cite{chan2022efficient} and NeRF-from-Image~\cite{pavllo2023shape} build NeRFs or its variants upon the outputs of a style-GAN~\cite{karras2019style}, while more
recent works~\cite{watson2022novel,pavllo2023shape,melas2023realfusion,xu2023neurallift,liu2023zero,chen2023single,gu2023nerfdiff,tang2023make,qian2023magic123,liu2023one,liu2023syncdreamer,chan2023generative} are favoring diffusion models such as Stable Diffusion (SD)~\cite{rombach2022high}. 
To date, Zero-1-to-3~\cite{liu2023zero} represents the state of the art on single- to multi-view synthesis; it is built on SD and fine-tuned by 800K 3D models from Objaverse~\cite{deitke2023objaverse}.

To alleviate the multi-view consistency problem, which is an reoccurring 
challenge for the image-space generative methods,
recent methods~\cite{qian2023magic123,liu2023syncdreamer,zhao2023efficientdreamer,long2023wonder3d,tang2023mvdiffusion} aim to enhance the consistency by performing spatial attention between different views, which dramatically increases the computational overhead.
Nevertheless, the consistency of two disparate views is still far from satisfactory. A deeper reason is that standard convolutional neural networks (CNNs) are not inherently rotation-invariant, making it difficult to build correspondences between significantly different views.





\vspace{2pt}

\mypara{Slice representation}
There have been techniques to use object slices to reconstruct a 3D shape, e.g., as inspired by Magnetic Resonance Imaging (MRI). 
Early works in computer graphics employ either parallel cross-sectional curves, e.g.,~\cite{kels2011cag}, or non-parallel curve networks, e.g.,~\cite{liu2008cgf}, as inputs.
Recently, with the advent of machine learning, OReX~\cite{sawdayee2023orex} and Cut-and-Approximate~\cite{ostonov2022cut} develop neural reconstruction models to
obtain 3D meshes, also from a sparse collection of 2D planar cross-sections, while NeuralSlice~\cite{jiang2023neuralslice} uses slicing to obtain 3D shapes from 4D tetrahedral meshes. 
However, none of these techniques, nor others that we are aware of, address the problem of single-view 3D reconstruction since their inputs all consist of 3D data such as 
curve slices or point clouds. Moreover, our method learns to predict much {\em fewer\/} slice {\em images\/} and use them as input to a transformer-based reconstruction module.

\vspace{2pt}

\mypara{Dip transform}
Dip transform~\cite{aberman2017dip} is an ingenious way of acquiring occluded structures of a 3D object by measuring fluid displacement when
the object is repeatedly oriented and dipped into the fluid. Similar to slicing, their method was also designed to ``reach'' occluded parts that optical 
captures via lines of sight cannot. However, both the approaches to occlusion revelation (liquid penetration vs.~slicing) and the target applications 
(3D acquisition vs.~single-view 3D reconstruction) are completely different.

\vspace{2pt}

\mypara{Multi-plane image (MPI)}
Introduced by Zhou et al.~\cite{zhou2018stereo}, an MPI representation for a scene consists of a set of fronto-parallel planes at fixed depths, 
with each image intended to capture scene appearance and visibility at a specific depth. Conceptually, both MPIs
and our slice images provide a layered scene representation for occlusion revelation. However,
MPIs~\cite{zhou2018stereo,flynn2019deepview,tucker2020single,han2022single,zhang2023structural} have only been employed for novel view synthesis, 
not 3D reconstruction, and the prediction networks were trained on {\em multi-view images\/},
e.g., from Youtube videos~\cite{zhou2018stereo}. The generated MPIs only reveal a small amount of occluded scenes to account for 
view shifts.
In contrast, our method directly cuts through a 3D object to reveal all the occluded parts at several depth ranges.

%% file: 02_method.tex
\section{Method}
\label{sec:method}

Our method is based on a slicing representation; see Section~\ref{sec:method_slice_rep} and Fig.~\ref{fig:slicer_rep}.
Given a single-view RGB image $I \in \mathbb{R}^{H \times W \times 3}$ of a 3D object, we first learn a slicing network (Section~\ref{sec:method_gen_slices}) which takes $I$ as input and outputs its corresponding multi-slice images.
Then, an SDF prediction network is trained (Section~\ref{sec:method_learn_imp_field}) to leverage image features from the slices to produce an implicit field of the object.

\subsection{Slicing Representation}
\label{sec:method_slice_rep}

As shown in Fig.~\ref{fig:slicer_rep}, given a 3D object {\em mesh\/}, we slice it along the $X$, $Y$, and $Z$ axes of its bounding box and render the sliced segments. 
The slice images can optionally include textures or have holes filled.
By default, we remove the textures and keep the holes, which could reveal back faces of the mesh, and overall, it produces the best experimental results. 
The rendered slice images consist of $\{S^{j}_{i} | 1 \le i \le N_{s}, j \in \{ X, Y, Z\}\}$, where $S^{j}_{i} \in \mathbb{R}^{H \times W \times 1}$, $j$ denotes the axis and $N_{s}$ is the number of slices along each axis. 
In our current implementation, we produce $N_{s} = 4$ slices per direction.
Compared to planar cross-sections or tomography scans which consist of densely stacked planar images, our slicing representation is more compact with each slice image revealing more object semantics.

\subsection{Slice Regression and Generation}
\label{sec:method_gen_slices}

Given an image $I$, observing a 3D shape from a single view, 
we aim to train a slicing network $F_{s}$ that can produce the slice images from $I$: $F_{s} (I) = \{\hat{S}_{i}^{j}\} $. 
We have two different designs for such a network: regressive and generative. 
The former aims to regress GT slices, assuming deterministic occluded parts based on the partial observation of the input view. The latter aims to \textit{generate} slices that are consistent with the input view, considering various possible occluded parts that do not necessarily match the GT slices. 


\subsubsection{Regression-based Slicing}

We employ a U-Net~\cite{ronneberger2015u} with an encoder and a decoder to regress the slices from $I$.
As shown in the top-left of Fig.~\ref{fig:pipeline}, 
an indicator code $E_{i}^{j} \in \mathbb{R}^{N_{e}}$ is concatenated with the latent code of $I$ to indicate which slice to output:

\begin{equation}
\hat{S}_{i}^{j} =  F_{d}([F_{e}(I); E_{i}^{j}]),
\end{equation}
where $F_{e}$ and $F_{d}$ are the encoder and decoder of $F_{s}$, respectively, and $[;]$ denotes concatenation.

The indicator codes $\{E^{j}_{i}\}$ are randomly initialized and jointly optimized with the whole network.
The loss function $\mathcal{L}_{reg}$ is the reconstruction error measured by $\mathcal{L}_{1}$-Norm and perceptual loss~\cite{zhang2018unreasonable} ($\mathcal{L}_{p}$) for all slices:
\begin{equation}
\mathcal{L}_{reg} = \frac{\sum_{i,j} || \hat{S}_{i}^{j} - S_{i}^{j}||_1 + \mathcal{L}_{p}(\hat{S}_{i}^{j}, S_{i}^{j})}{|S|},
\end{equation}
where $|S| = 3 \cdot N_{s}$ is the total number of slices. 

\subsubsection{Generation-based Slicing}

\input{figs/fig_diff_process}

Single-view occlusion revelation is clearly ill-posed, with potential ambiguities leading to 
multiple plausible 3D reconstructions. To this end, 
we consider slicing as an image-conditioned generation task and develop a generation-based slicing via Denoising Diffusion Probabilistic Models (DDPM)~\cite{ho2020denoising}, whose detailed pipeline is shown in Fig.~\ref{fig:diff_process}.

The key challenge of applying DDPM lies in how to generate {\em consistent\/} slices.
Rather than diffusing each single slice, we propose to concatenate all slices into an aggregated structure $[S]$, along either the color dimension ($[S] \in \mathbb{R}^{H \times W \times |S|}$) or a spatial dimension such as height ($[S] \in \mathbb{R}^{(|S| \cdot H) \times W \times 1}$).
The concatenation of slices instructs DDPM to learn the correlation of slices, yielding consistent generation results.
By default, we concatenate along the color dimension.
The comparison of these two concatenation ways can be found in the supplementary.
\par

Following the procedures of DDPM, we first inject a noise $\epsilon \sim \mathcal{N}(0, 1)$ to $[S]$ to obtain $[\bar{S}]$. The noised slices $[\bar{S}]$ are then sent to a U-Net based denoising network $F_{dn}$ that outputs the predicted noise conditioned on $I$: 

\begin{equation}
\hat{\epsilon} = F_{dn}([\bar{S}], I).
\end{equation}

In practice, the injection and prediction of noise $\epsilon$ involve an iteration step $t$, which we omit here for brevity. 
To better condition $I$ for denoising diffusion, we employ a pre-trained VGG to extract multi-layer features from $I$ and add them into the corresponding layers in the encoder of $F_{dn}$ (see the blue part of Fig.~\ref{fig:diff_process}). The loss function is the denosing loss: 
$\mathcal{L}_{gen} = || \hat{\epsilon} - \epsilon ||^2_{2}.$
After training, we can sample random noises to generate a set of slice images $\{\hat{S}_{i}^{j}\}$.

Compared to regression-based slicing, generation-based slicing can produce multiple plausible results rather than a single deterministic result, but consumes much longer inference time due to the nature of diffusion models. 

\subsection{Learning Implicit Field from Slices}
\label{sec:method_learn_imp_field}

We learn an implicit field from slice images, by projecting a 3D query point onto each slice image, extracting query-specific features, and then aggregating them to predict a signed distance. For regression-based slicing, we train the SDF prediction network on the predicted slices from the previous section thus the whole pipeline can be trained end-to-end. For generation-based slicing, we train the SDF prediction network on GT slice images since the generated slices may not match the GT signed distances. 

Specifically, for a query point $q \in \mathbb{R}^{3}$, we project it to a location $\tilde{q} 
\in \mathbb{R}^{2}$ on the 2D image plane according to the camera parameters $\Theta$. 
For slice $\hat{S}^{j}_{i}$ (regression) or $S^{j}_{i}$ (generation),
we first extract its feature maps, then retrieve features from each feature
map corresponding to location $\tilde{q}$, and finally concatenate them to obtain the image feature vector $V^{j}_{i} \in \mathbb{R}^{1 \times 1 \times N_c}$. 
The retrieval is accomplished via bi-linear interpolation over the features grids that encompass the location of $\tilde{q}$.
We design a transformer network~\cite{vaswani2017attention} $F_{t}$ which learns to aggregate the feature vectors from all slices to predict the signed distance $\hat{d}$: 
$\hat{d} = F_{t}([q^{\prime}, V^{X}_{1}, ..., V^{Z}_{N_{s}} ]),$
where $q^{\prime} \in \mathbb{R}^{N_c}$ is linearly projected from $q$ to be the same dimension with $V^{j}_{i}$.
The optimization objective is to minimize the $\mathcal{L}_{1}$ distance between predicted and GT signed distance: $\mathcal{L}_{imp} = || \hat{d} - d ||_1$. In regression-based slicing, $\mathcal{L}_{imp}$ and $\mathcal{L}_{reg}$ are jointly optimized.



As in DISN~\cite{xu2019disn}, we extract from only the input view $I$ to obtain a feature vector $V$. For query points located around the occluded parts of the object (e.g., the query point in Fig.~\ref{fig:pipeline}), the feature $V$ is irrelevant to $q$ since it is extracted from the front parts that block $q$, often leading to erroneous SDF predictions.
In contrast, one or more of our slice images could reveal the occluded parts, yielding accurate features fed to the transformer to predict the correct SDF. 
After learning the implicit field, we apply the Marching Cube algorithm~\cite{lorensen1998marching} to obtain triangle meshes.

\subsection{Camera Pose Estimation}
\label{sec:method_cmr_est}
We follow the strategy in DISN~\cite{xu2019disn} to estimate the camera poses. 
First, assuming a fixed set of intrinsic parameters, we only need to predict a translation vector $\Theta_{t} \in \mathbb{R}^{3}$ and a rotation matrix $\Theta_{r} \in \mathbb{R}^{3 \times 3}$.
Then, a CNN, e.g., VGG-16~\cite{SimonyanZ14a}, is trained to estimate $\hat{\Theta}_{t}$ and $\hat{\Theta}_{r}$ from input view $I$.
Afterwards, we sample a point cloud $P \in \mathbb{R}^{N_p \times 3}$ from the object along with its camera-aligned version $P'$.
Finally, the loss function of the CNN is to align $P'$ with $P$ using $\hat{\Theta}_{t}$ and $\hat{\Theta}_{r}$, i.e., $\mathcal{L}_{cam} = \frac{1}{N_p}|| P - (P'\Theta_{r} + \Theta_{t}) ||^2_{2}$.

%% file: figs/fig_diff_process.tex
\begin{figure}[!t]
    \centering
    \includegraphics[width=0.98\columnwidth]{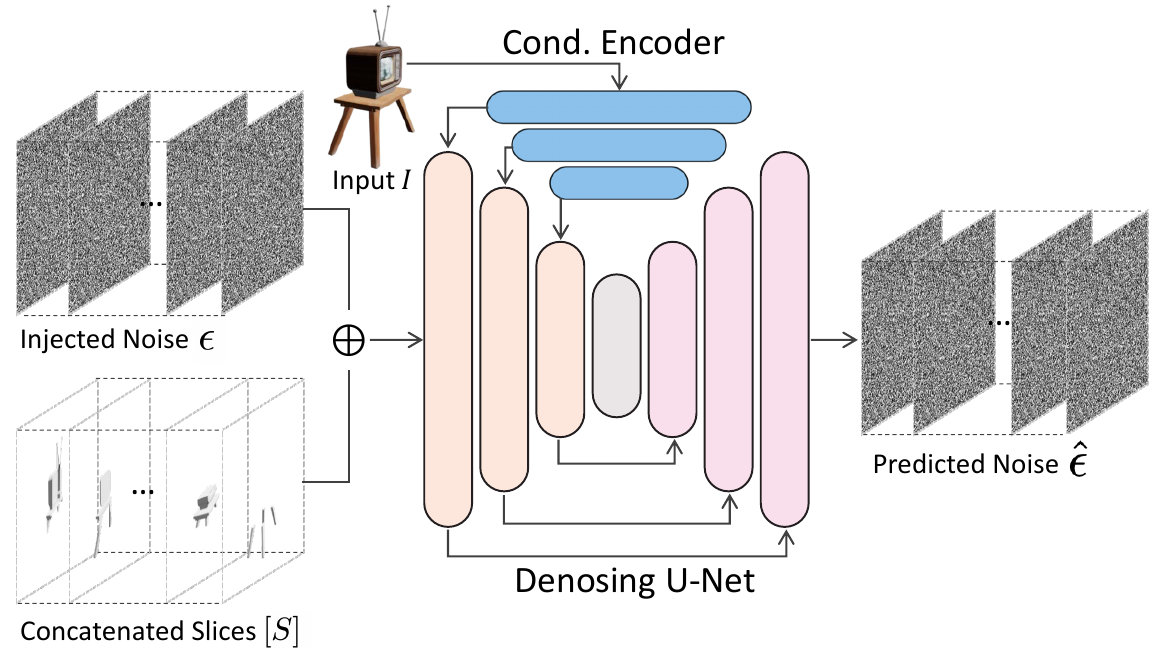}
\caption{The pipeline of generation-based slicing via DDPM. $\bigoplus$ denotes the injection of noise to slice images.
}
\label{fig:diff_process}
\end{figure}

%% file: 03_results.tex
\section{Results and Evaluation}
\label{sec:results}


\mypara{Dataset} 
We utilize two well-known datasets, ShapeNet~\cite{chang2015shapenet} and Objaverse~\cite{deitke2023objaverse}, to train and test our network.
For ShapeNet, our method is applied to Chairs and Airplanes separately, using the train/validation/test split as in SHREC16~\cite{savva2016shrec16}. 
Objaverse is a challenging dataset, 
where many objects may not belong to a clear category.
In our experiment, we randomly selected $\approx$40K shapes from Objaverse to keep resource and time costs manageable, while ensuring diversity of the collection.
We follow a 95\%-5\% train-test split over the 40K selected shapes, ensuring that every training model we use is also from the training set of Zero-1-2-3~\cite{liu2023zero}.
Note that Zero-1-2-3 used a train-test split of 99\%-1\% over the entire 800K models in Objaverse.


\mypara{Rendering Slice Images}
We employ the \textit{bisect} function in Blender~\cite{blender_bisect} to cut an object mesh into slices and then render them.
By default, we keep internal cavities and holes produced by slicing.
Since many 3D models in Objaverse are not in, or have no, canonical poses and feature random rotations, slicing them along the default $X$, $Y$, and $Z$ directions is not suitable.
To this end, we opted for \emph{camera-aligned} slicing instead. Specifically, we set the camera position, introduce a random rotation to the object, position it in front of the camera, and utilize the camera's axes for slicing. 
Under this setting, estimating the rotation of objects is unnecessary; see more details in the supplementary material.




\mypara{Implementation Details}
When preparing training data, we use the Vega FEM Library~\cite{sin2013vega} for SDF computation, same as DISN~\cite{xu2019disn} and AutoSDF~\cite{mittal2022autosdf}.
For each object, the diagonal length of the object's 3D bounding box is normalized to 1. We set $H$ and $W$ to $128$, $N_{e}$ to $128$, $N_{s}$ to $4$, and $N_{p}$ to 1,024. 
We train our model on {\em only one\/} Nvidia A40 GPU. 

\mypara{Evaluation Metrics} We use Chamfer $\mathcal{L}_{2}$ Distance (CD), F-score\%1 (F1)~\cite{tatarchenko2019single}, and Hausdorff Distance (HD) to evaluate reconstruction results.
Besides these 3D metrics, we also evaluate using 2D metrics, PSNR, SSIM~\cite{wang2004image}, and LPIPS~\cite{zhang2018unreasonable}, on 24 rendered views of reconstructed meshes.

\subsection{Comparisons to State Of The Art (SOTA)}
We compare our method with SOTA methods including DISN~\cite{xu2019disn}, AutoSDF~\cite{mittal2022autosdf}, NeRF-From-Img~\cite{pavllo2023shape}, SSDNeRF~\cite{chen2023single}, RealFusion~\cite{melas2023realfusion}, and One-2-3-45~\cite{liu2023one}. 
Note that when showing visual results, we always present two views to remove any view bias towards the input.



\input{tables/tab_comp_quan_shapenet_light}
\input{figs/fig_comp_qual_sp_chairs}
\input{figs/fig_comp_qual_sp_others}

\mypara{ShapeNet} Figs.~\ref{fig:comp_qual_sp_chairs} and~\ref{fig:comp_qual_sp_others} demonstrate the effectiveness of our approach in reconstructing obscured sections of objects while precisely restoring visible ones.
In contrast, DISN~\cite{xu2019disn} can barely reconstruct the occluded parts.
AutoSDF~\cite{mittal2022autosdf} and NeRF-From-Image~\cite{pavllo2023shape} rely on latent features for reconstruction, which can yield rather clean results but they do not necessarily respect the input image, as the chair
results from Fig.~\ref{fig:comp_qual_sp_chairs} evidently showed.
SSDNeRF~\cite{chen2023single} tends to produce erroneous geometric details and the surfaces are not as smooth as from other methods, especially for the airplane examples in Fig.~\ref{fig:comp_qual_sp_others}.
Generally speaking, One-2-3-45~\cite{liu2023one} yields rough and bulky geometries that clearly lack details. 
Additionally, our method outperforms the other methods in the reconstruction of unconventional inputs, as shown by the two airplanes in Fig.~\ref{fig:comp_qual_sp_others}.

Tab.~\ref{tab:comp_quan_shapenet_light} shows quantitative comparisons to the above methods on the two ShapeNet categories.
When our method operates without using GT camera poses and relies on estimated poses, its performance sees only a marginal degradation.
Notably, for Chair, our method with estimated poses even outperforms DISN and SSDNerf with GT poses.
It is also notable that One-2-3-45 tends to output dilated shapes. Although it falls behind DISN in quantitative results, it usually generates more complete shapes and more accurate topologies than DISN; see Figs.~\ref{fig:comp_qual_sp_chairs} and \ref{fig:comp_qual_sp_others}.

\mypara{Objaverse}
\input{tables/tab_comp_quan_objaverse}
\input{figs/fig_comp_qual_objaverse}
Since AutoSDF~\cite{mittal2022autosdf}, NeRF-From-Image~\cite{pavllo2023shape}, and SSDNeRF~\cite{chen2023single} are all category-specific methods, they cannot be trained on Objaverse which consists of many 3D objects with no clear categories. 
Thus, we mainly compare our method with DISN~\cite{xu2019disn},  RealFusion~\cite{melas2023realfusion}, and One-2-3-45~\cite{liu2023one}; see Fig.~\ref{fig:comp_qual_objaverse} for a qualitative comparison.

Note that RealFusion~\cite{melas2023realfusion} and One-2-3-45~\cite{liu2023one} both leverage a significantly larger volume of training data compared to our method, as they both rely on a pre-trained Stable Diffusion~\cite{rombach2022high} model.
Despite this disparity in the scale of the training sets, our method still outperforms these two methods, as well as DISN; see Fig.~\ref{fig:comp_qual_objaverse}.
Quantitatively, Tab.~\ref{tab:comp_quan_objaverse} demonstrates our method's across-the-board advantage over DISN and One-2-3-45 (more so on Objaverse than on ShapeNet), in both 3D and 2D metrics, especially CD, HD, and LPIPS.
These results also show that our method possesses the capacity to generalize across common shapes rather than being confined to specific categories.


\mypara{Real-world Image Input}
To further stress-test the methods, we perform Google image search on object keywords (e.g., ``chair'', ``airplane", etc.) and take the top-1 image with occlusions as input. In addition, we use
some examples from Google Scanned Objects (GSO)~\cite{downs2022google} dataset.
Fig.~\ref{fig:comp_qual_real_data} shows that our method performs the best in recovering occluded parts while preserving fine details, e.g., the turbines of the airplane, the heel, and the opening of the boot.

\input{figs/fig_comp_qual_real_data}

\input{figs/fig_generative_slices}

\mypara{Multi-slice vs.~Multi-view Reconstruction}
Fig.~\ref{fig:generative_slices} compares between multi-slice and multi-view reconstructions.
We demonstrate two different sets of multi-slices and multi-views generated from our method and One-2-3-45~\cite{liu2023one}.
Our generated slice images depict two distinct configurations of chair legs, maintaining high consistency across slices. In contrast, the multiple views synthesized in One-2-3-45 do encounter inconsistency issues, compromising an accurate 3D reconstruction.
We show three slices in each set and more can be found in the supplementary material.


\mypara{Training and Inference Costs}
Unlike RealFusion~\cite{melas2023realfusion} and One-2-3-45~\cite{liu2023one}, our method does not depend on a large pre-trained model~\cite{rombach2022high}. NeRF-based methods tend to have long inference times, as many of them require per-test optimization. 
Our method has much faster inference times: \textbf{10 or 20 seconds} with slice regression or generation, compared to 90 minutes for RealFusion and 45 seconds for One-2-3-45 and SSDNeRF, while delivering superior results. With similar inference times to DISN, Auto-SDF, and NeRF-From-Img, our method has better reconstruction quality. See supplementary material for more details.

\subsection{Ablation Study}
\label{sec:exp_ablation}

\mypara{Generative vs.~Regressive Slicing}
As shown in the first row of Fig.~\ref{fig:generative_slices}, for generation-based slicing, we can produce multiple plausible results for a given input view. 
Tab.~\ref{tab:comp_quan_objaverse} shows that regressive slicing outperforms generative slicing when compared to GT. However, it is notable that the 3D outputs may differ from GT when input views have ambiguous occluded parts.
Regressive training struggles to handle such one-to-many mappings, leading to generation outperforming regression in reconstructing occluded parts.

\mypara{Number of Slices}
We test the performance of our method with regression-based slicing by varying $N_{s}$ between 2, 4, and 8, 
on ShapeNet Chairs.
From Tab.~\ref{tab:param_study_and_abaltion}, it is evident that larger $N_{s}$ leads to better reconstruction quality.
However, $N_{s} = 8$ will dramatically increase the computational cost of our method since its complexity is $O(N_{s}^{2})$, as pair-wised attentions are performed on the slices in the transformer.
As a trade-off between reconstruction quality and computational complexity, we set $N_{s} = 4$ by default.

Tab.~\ref{tab:param_study_and_abaltion} also covers several other ablation studies.

\mypara{Learned v.s.~One-hot Slice Indicator}
Learned indicator codes (the default setting) achieve slightly better performance than using one-hot codes to indicate slice locations. This can be seen by
comparing numbers from the first half row of Tab.~\ref{tab:param_study_and_abaltion} to those in the last row of Tab.~\ref{tab:comp_quan_shapenet_light}.

\input{tables/tab_abl_num_slices_only_3d}

\mypara{Slices v.s.~Depth Images}
Since our slice images offer layered information, it is logical to perform a comparison against depth images as supplementary information.
We compared the performance of DISN when concatenating estimated depth images with the input view. Results are shown in the ``DISN+Depth'' half row, which indicates that depth images can only provide marginal improvement.


\mypara{Three Axes v.s.~One Axis} 
We also investigated the option of slicing along only one axis. Not surprisingly, Tab.~\ref{tab:param_study_and_abaltion} shows that this underperforms against using three axes.

\mypara{Solid v.s~Hollow Slices}
Tab.~\ref{tab:param_study_and_abaltion} also demonstrates that not filling the holes improves the performance as more information about the inner/outer surfaces is revealed this way.


\mypara{Using Textured Slices}
We can optionally keep object textures in the slice images.
Experimental results show that removing the textures in the slice images can achieve better reconstruction. The reasons could be: 1) learning to produce slice images without textures could help the model better distinguish geometries from textures; 2) for some 3D models the color of inner and outer surfaces could be different which may complicate the learning process.

%% file: tables/tab_comp_quan_shapenet_light.tex
\begin{table}[!t]
    \resizebox{0.98\columnwidth}{!}{
    \begin{tabular}{ccccccc}
    \toprule
     Datasets               & \multicolumn{3}{c}{ShapeNet Chairs} & \multicolumn{3}{c}{ShapeNet Airplanes} \\ 
     
     Metrics               & CD$\downarrow$  & F1$\uparrow$ & HD$\downarrow$ &  CD$\downarrow$  & F1$\uparrow$ & HD$\downarrow$  \\ 
     \midrule
     DISN~\cite{xu2019disn}                 & 3.25 &  2.34 & 7.84 & 3.48 & 2.81 & 7.62  \\
     Auto-SDF~\cite{mittal2022autosdf}      & 25.0 & 0.77 & 15.8 & - & - & - \\
     NeRF-Img~\cite{pavllo2023shape} & 19.5 & 1.27 & 13.4 & 16.0 & 1.41 & 14.0 \\

     One-2-3-45~\cite{liu2023one} & 12.5 & 1.01 & 11.7 & 23.4 & 0.75 & 12.6\\
     Ours & 2.80 & 2.43 & 6.75 & 2.99 & 2.83 & 6.45 \\
      \midrule
     DISN*~\cite{xu2019disn} & 3.02 & 2.42 & 7.86 & 3.19 & \textbf{2.96}& 7.56\\

     SSDNeRF*~\cite{chen2023single} & 8.98 & 2.06 & 10.9 & 10.4 & 2.00  & 10.4    \\
     Ours* &  \textbf{2.67} & \textbf{2.53} & \textbf{6.70}   &   \textbf{2.81} & \textbf{2.96} & \textbf{6.38}  \\
     \bottomrule
\end{tabular}}
     \vspace{-0.2cm}
     \caption{Quantitative results of single-view 3D reconstruction on ShapeNet Chairs and Airplanes. `*' denotes using ground-truth (GT) camera poses. Slice3D (ours) uses regression-based slicing.} 
    \label{tab:comp_quan_shapenet_light}
\end{table}

%% file: figs/fig_comp_qual_sp_chairs.tex
\begin{figure*}[!t]
    \centering
    \includegraphics[width=0.98\textwidth]{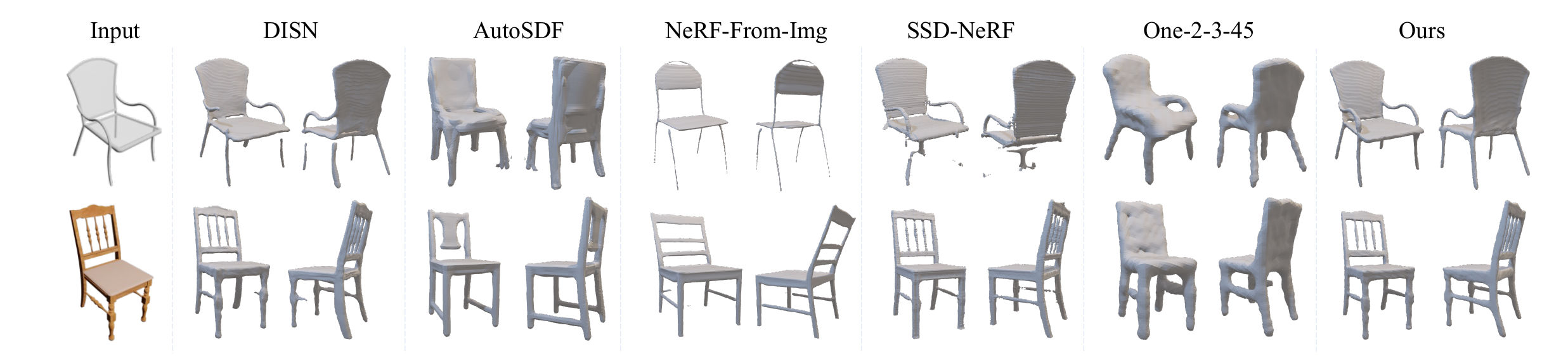}
\caption{Visual comparison between single-view 3D reconstruction methods on two ShapeNet chairs. DISN and our method (based on regressive slicing) utilize the same estimated camera parameters. Two different views are displayed to remove view bias.}
\label{fig:comp_qual_sp_chairs}
\end{figure*}

%% file: figs/fig_comp_qual_sp_others.tex
\begin{figure*}[!t]
    \centering
    \includegraphics[width=0.98\textwidth]{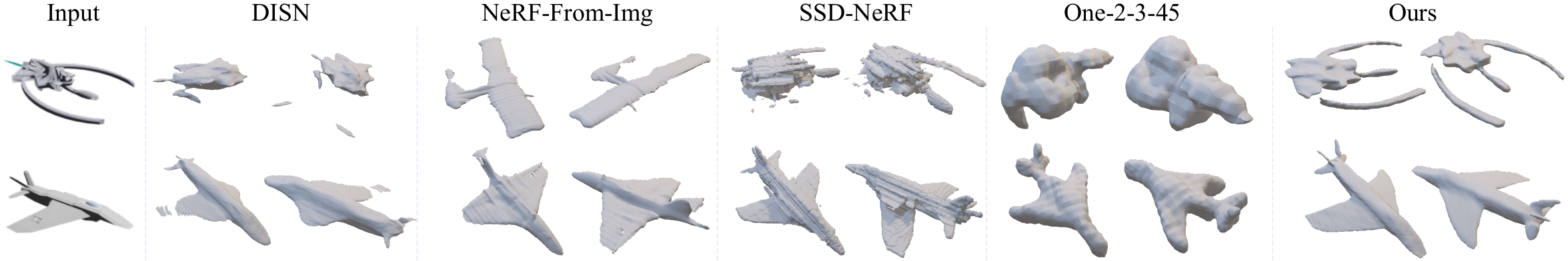}
\caption{Visual comparison between single-view 3D reconstruction methods on two somewhat unconventional airplanes from ShapeNet.
}
\label{fig:comp_qual_sp_others}
\end{figure*}

%% file: tables/tab_comp_quan_objaverse.tex
\begin{table}[t]
    \resizebox{1.0\columnwidth}{!}{
    \begin{tabular}{ccccccc}
     \toprule
     Method               & CD$\downarrow$  & F1$\uparrow$ & HD$\downarrow$ & PSNR$\uparrow$ & SSIM$\uparrow$ & LPIPS$\downarrow$  \\ \midrule
     \addlinespace[4pt] 
     DISN~\cite{xu2019disn}                 & 28.2  & 1.47 & 23.0 & 21.9 & 8.79 & 15.2  \\
     One-2-3-45~\cite{liu2023one}      & 35.4 & 0.87 & 29.8 & 19.8 & 8.55 & 19.8 \\
     Ours (G) & 25.0 & 1.51 & 16.4 & 22.3 & 8.84 & 13.4 \\
     Ours (R) & \textbf{16.6} & \textbf{1.53} & \textbf{13.4} & \textbf{22.6} & \textbf{8.88} & \textbf{12.8} \\
     \bottomrule
\end{tabular}}
     \vspace{-0.2cm}
     \caption{Quantitative results of single-view 3D reconstruction on the Objaverse dataset. `R' and and `G' denote regression-based and generation-based slicing in Slice3D, respectively.}
    
    \label{tab:comp_quan_objaverse}
\end{table}

%% file: figs/fig_comp_qual_objaverse.tex
\begin{figure*}[!t]
    \centering
    \includegraphics[width=0.98\textwidth]{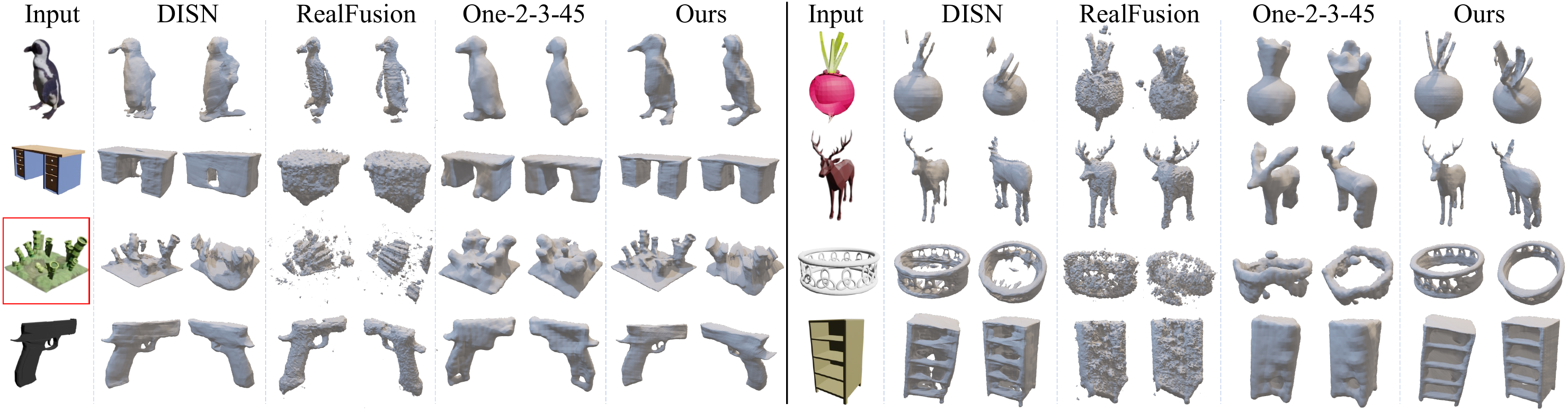}
\caption{Visual comparison between cross-category reconstruction methods on Objaverse. Our method better recovers overall geometries, fine details, and is more faithful to the inputs. Except for the seaweed (squared), all the other models are in the {\em training\/} set of Zero-1-to-3.}
\label{fig:comp_qual_objaverse}
\end{figure*}

%% file: figs/fig_comp_qual_real_data.tex
\begin{figure}[!t]
    \centering
    \includegraphics[width=0.97\columnwidth]{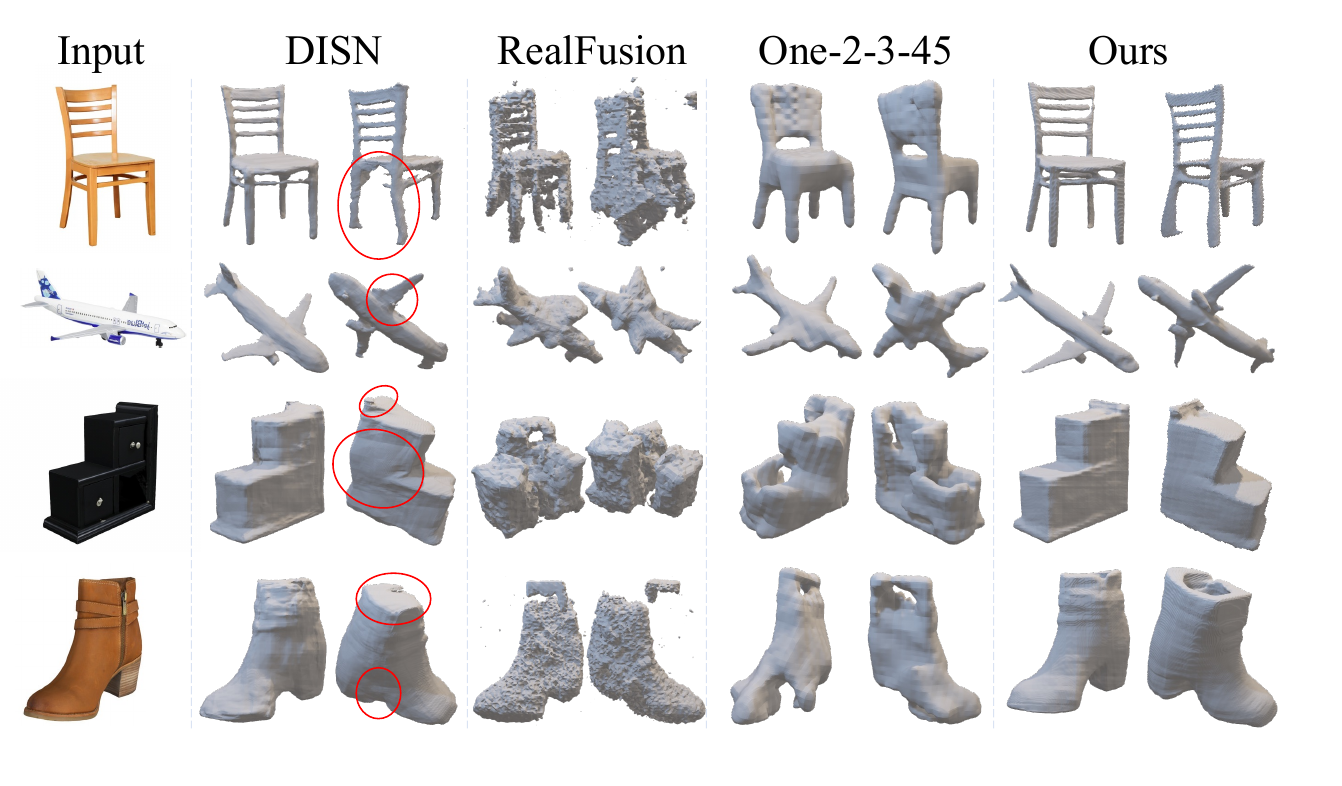}
\caption{On real-world image inputs. Top two from Google image search and the others from GSO~\cite{downs2022google}. 
Zoom in for details.
}
\label{fig:comp_qual_real_data}
\end{figure}

%% file: figs/fig_generative_slices.tex
\begin{figure*}[!t]
    \centering
    \includegraphics[width=0.98\textwidth]{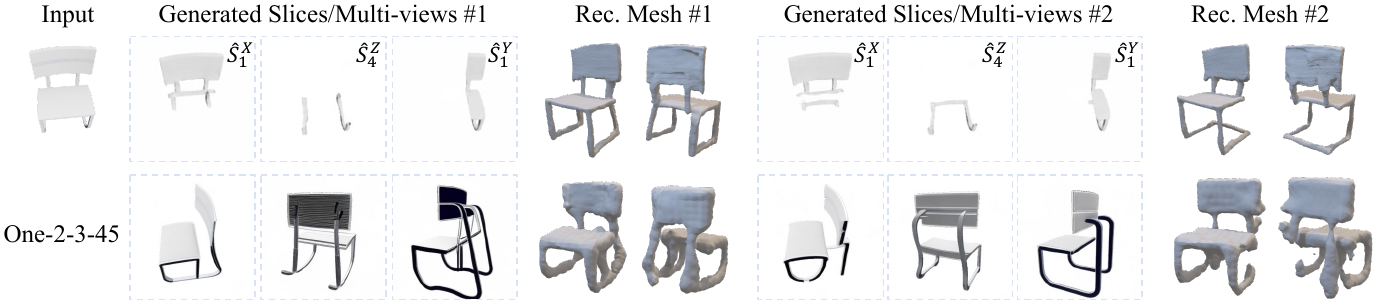}
\caption{Multi-slice vs.~multi-view reconstructions amid ambiguities in the chair legs. Both One-2-3-45 (bottom) and Slice3D (top) can produce multiple results. Our results are both plausible from consistent slices, while One-2-3-45
suffers from multi-view inconsistencies.
}
\vspace{-0.2cm}
\label{fig:generative_slices}
\end{figure*}

%% file: tables/tab_abl_num_slices_only_3d.tex
\begin{table}[t]
    \resizebox{0.98\columnwidth}{!}{
    \begin{tabular}{lrrrlrrr}
    \toprule
     Setting               & CD$\downarrow$  & F1$\uparrow$ & HD$\downarrow$ &  Setting               & CD$\downarrow$  & F1$\uparrow$ & HD$\downarrow$  \\ 
     \midrule
     One-hot ind. & 2.68  & 2.53 & 6.82 &  only $X$ & 2.88 & 2.39 & 7.74 \\
     DISN+Depth &  2.99 & 2.43 & 7.81 &  only $Y$ & 2.75 &2.45  & 6.84 \\
     $N_s = 2$   & 2.90 & 2.46 & 7.31 &  only $Z$ & 3.08 & 2.38 & 7.60   \\
     $N_s = 4$   &   2.67 & 2.53 & 6.70  &  Filled & 2.73 & 2.50 & 7.37 \\
     $N_s = 8$ & 2.56 & 2.56 & 6.93 &  w/ textures & 2.82  & 2.50 & 7.70 \\
     \bottomrule
\end{tabular}}
     \vspace{-0.2cm}
     \caption{Ablation studies on ShapeNet Chairs.} 
    \label{tab:param_study_and_abaltion}
\end{table}

%% file: 04_conclusion.tex
\section{Discussion and Future Work}
\label{sec:future}


Our multi-slice idea represents a significant departure from conventional approaches to single-view 3D reconstruction. Through conceptual illustrations (Fig.~\ref{fig:rec_interiors}) and extensive experiments, we
demonstrate its compelling properties in recovering occluded object structures. Note again that all the Slice3D results on Objaverse~\cite{deitke2023objaverse} were obtained by networks trained on a very {\em small subset\/}
(40K or \textbf{5\%}) of the 800K 3D models therein, whereas One-2-3-45~\cite{liu2023one} utilizes Zero-1-to-3~\cite{liu2023zero} for multi-view synthesis and the latter was
fine-tuned on the entirety of the 800K. Yet our results (Table~\ref{tab:comp_quan_objaverse}) show that Slice3D outperforms One-2-3-45 across the board, suggesting that our model generalizes better,
likely owning to its superior occlusion-revealing abilities.

Our current implementation only uses four slices per direction, which is quite coarse.
Since there is no camera rotation, parts of an object far away from the camera are in smaller scale, e.g., the rear leg of the chair in Fig.~\ref{fig:comp_qual_real_data} and of the buck in Fig.~\ref{fig:comp_qual_objaverse}. This can hinder our network's ability to extract detailed features. More refined and gradual slicing over those parts may be needed to improve matters.

The foremost limitation of slicing however, is that realistically, it can only be executed at scale on {\em digital\/} 3D models, not physical ones. On the other hand, the multi-view approach can be applied to physical 3D objects by capturing 
photos only, without ever needing 3D supervision, as in RealFusion~\cite{melas2023realfusion}. However, this is not without a high cost: 90 minutes by RealFusion vs.~45 seconds by One-2-3-45 (requiring 3D fine-tuning
for multi-view synthesis) and less than 20 seconds by Slice3D. We envision synthesizing 3D models, not many but carefully crafted few, for the multi-slice approach as an intriguing direction to explore.

%% file: 05_supp.tex
\clearpage

\maketitlesupplementary

\section{Rendering Details}
\label{sec:render_details}
Given a 3D model, we first normalize the length of its body diagonal to 1.
Then, we put the 3D model in front of the camera with a distance of 1.2, i.e., the camera is pointed to the centre of the bounding box of the object with a distance of 1.2.
Next, we randomly give a rotation, scaling and translation to the object.
The rotation is done by introducing a pair of elevation and azimuth randomly sampled from the ranges of [-10\degree, 40\degree] and [0\degree, 360\degree], respectively.
When introducing scaling and translation, we restrict the 3D model to be always inside $[-0.5,0.5]^{3}$, from where the query points are sampled in both training and inference. 

This rendering rule is implemented in both ShapeNet~\cite{chang2015shapenet} and Objaverse~\cite{deitke2023objaverse}. 
Note that positions of an object and a camera are relative. If the object is assumed to be fixed, we can tailor the rendering specifics by adjusting the camera through rotation and translation. In Section 3.4 of our main paper, we assume the object is fixed and then estimate rotations and translations of the camera.

\section{Visualization of Hole-Filled Slices}
\label{sec:hole-filled}

As our ablation studies already showed (Tab.~\ref{tab:param_study_and_abaltion}), operating on slice images {\em without\/} the holes filled provides more information about both the inside and outside structures of an object, yielding better reconstruction quality.
In Fig.~\ref{fig:fill_vs_no_fill_slicings}, we visually compared slice images with (left) vs.~without holes filled.

\input{figs/fig_fill_vs_no_fill_slicings}

\section{Details on Two Slicing Directions}

As mentioned in Section 4, we utilize camera-aligned slicing for Objaverse dataset where many 3D models feature arbitrary orientations, i.e., they do not have clear canonical poses.
In other cases, e.g., for ShapeNet shapes, we rely on the canonical poses of the objects and perform slicing along the $X$, $Y$, and $Z$ directions with respect to those canonical poses.
Fig.~\ref{fig:supp_slicing_canonical_vs_camera} illustrates and contrasts these two choices for the slicing directions, using a chair example from ShapeNet dataset.

The 3D models in ShapeNet possess canonical poses (e.g., chairs consistently have a front orientation) and they are aligned along the default X, Y, and Z axes. Consequently, objects are always sliced along their default X, Y, and Z axes regardless of the rendered views.

For 3D models in Objaverse, we use camera-aligned slicing.
As described in Sec.~\ref{sec:render_details}, the camera is fixed and an object will be randomly rotated, scaled, and translated.
We slice the bounding boxes of the object along the axes in the camera world.

\section{Whether to Estimate Camera Poses}

For our method, camera pose estimation is necessary only when we want to output 3D shapes in their canonical poses.
In ShapeNet, we follow most existing methods~\cite{xu2019disn,mittal2022autosdf,chen2023single,pavllo2023shape} to produce 3D shapes in their canonical poses, which involves camera pose estimation.
However, in Objaverse, where many objects do not have canonical poses and exhibit random orientations, we avoid estimating the rotations and translations of objects, and reconstruct the 3D shapes as they are in the camera world.


\section{Training Cost and Inference Speed}
Tab.~\ref{tab:comp_train_cost_infer_speed} makes a detailed comparison of training cost and inference speed among single-view-
reconstruction methods.
Our method does not rely on big pre-trained model like Stable Diffusion~\cite{rombach2022high}.
Compared to multi-view based methods, 
our method runs much faster in the process of producing 3D meshes from slice images because we employ neural signed distance filed instead of NeRFs whose optimization is time-consuming.

\input{tables/tab_supp_comp_train_cost_infer_speed}
\input{figs/fig_supp_slicing_canonical_vs_camera}
\input{figs/fig_supp_comp_qual_sp_chairs_syncdreamer}
\input{figs/fig_supp_slice_visualization}

\section{Concatenation of Slice Images in Diffusion}
We perform DDPM~\cite{ho2020denoising} on the entirety of slice images that can be stacked either on the color dimension or a spatial dimension.
Note that concatenating along a spatial dimension significantly increases complexity because of the self attentions operated on the spatial dimensions in a diffusion network.
By default, we stack them on the color channel to reduce the training and inference time.
Tab.~\ref{tab:slices_concatenation_in_diffusion} provides a quantitative comparison of these two concatenation methods, revealing that
concatenating along a spatial dimension achieves better performance than the color dimension.
This outcome is logical as the former can model the spatial correspondence of different slice images throughout the diffusion network.
Given sufficient computational resources, prioritizing concatenation along a spatial dimension is recommended.

\input{tables/tab_supp_slices_concatenation_in_diffusion}

\section{View Inconsistency Problem}
As mentioned in our main paper, recent methods~\cite{qian2023magic123,liu2023syncdreamer,zhao2023efficientdreamer,long2023wonder3d,tang2023mvdiffusion} aim to enhance the consistency of synthesized views by performing spatial attention across different views.
In Fig.~\ref{fig:supp_comp_qual_sp_chairs_syncdreamer}, 
a comparison is made between Slice3D and SyncDreamer~\cite{liu2023syncdreamer} from these works. 
The findings indicate that despite the utilization of expensive spatial attentions, the challenge of maintaining view consistency persists. This further substantiates the advantages of employing multi-slices over multi-views.

\section{Visualization of Predicted Slice Images}

The predicted slice images for the examples in Fig.~\ref{fig:comp_qual_sp_chairs} and~\ref{fig:comp_qual_sp_others} can be found in Fig.~\ref{fig:supp_generative_slices_visualization} and Fig.~\ref{fig:supp_generative_slices_show_full}, respectively.
Notably, our slice3D can produce slice images with a high level of consistency.

\section{More Visual Results}

More visual results and comparisons are provided in Fig. ~\ref{fig:supp_comp_qual_sp_chairs} and~\ref{fig:supp_comp_qual_sp_cars} for ShapeNet, Fig.~\ref{fig:supp_comp_qual_objaverse} for Objaverse, and Fig.~\ref{fig:supp_comp_qual_gso} for Google Scanned Objects (GSO)~\cite{downs2022google}. As apparent, our results respect the geometric details better than the other techniques while they do not suffer from unwanted artifacts or noise. Also compared to other techniques such as AutoSDF, it better respects the input image and does not retrieve a model that might look clean and noise-free but it is far from the input image (e.g., Fig.\ref{fig:supp_comp_qual_sp_chairs}; first two rows).

\section{Image Resolution}

Due to limited computing resources, the resolution of our input images and slice images is set to only $128$. We plan to increase the resolution to $256$ or $512$ in the future and produce 3D meshes with better quality and details.

\input{figs/fig_supp_gen_slices_show_full}
\clearpage

\input{figs/fig_supp_comp_qual_sp_chairs}
\input{figs/fig_supp_comp_qual_sp_cars}
\input{figs/fig_supp_comp_qual_objaverse}
\input{figs/fig_supp_comp_qual_gso}

%% file: figs/fig_fill_vs_no_fill_slicings.tex
\begin{figure}[h]
    \centering
    \includegraphics[width=0.98\columnwidth]{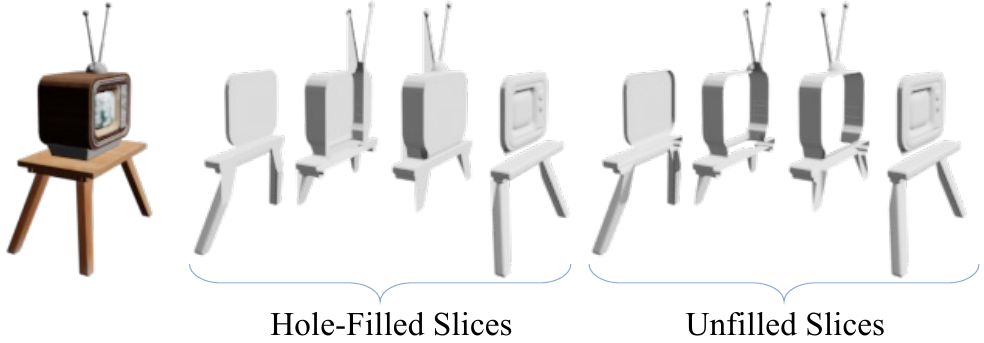}
\caption{Slicing with and without filling holes. The demonstrated images are sliced along axis X of the object.
}
\label{fig:fill_vs_no_fill_slicings}
\end{figure}

%% file: tables/tab_supp_comp_train_cost_infer_speed.tex
\begin{table}[h]
    \resizebox{1.0\columnwidth}{!}{
    \begin{tabular}{ccccc}
     \toprule
     Method     & Pre-trained model         & Training data  & Training GPU \& time &  Infer. speed  \\ \midrule
     \addlinespace[4pt] 
     DISN~\cite{xu2019disn}                 & VGG~\cite{SimonyanZ14a} & SPN-Chair &  1 $\times$ A40 for 1 day &  $<$5s  \\
     AutoSDF~\cite{mittal2022autosdf}                 & ResNet-18~\cite{he2016deep} & SPN-Chair & unknown &  $<$10s  \\
     NeRF-Img~\cite{pavllo2023shape}                 & VGG-16~\cite{SimonyanZ14a} & SPN-Chair & 4 $\times$ A100 &  $<$30s  \\
     SSDNeRF~\cite{chen2023single}                 & N/A & SPN-Chair & 2$\times$ 3090 for 6 days & 45s-1min   \\
     Ours & VGG-16~\cite{SimonyanZ14a} & SPN-Chair & 1 $\times$ A40 for 2 days &  $<$10s(R)/$<$20s(G) \\
     \midrule
     Real-Fusion~\cite{melas2023realfusion}                 & SD~\cite{rombach2022high} & Per-case opt. & N/A &  ~90min   \\
     \midrule
     One-2-3-45~\cite{liu2023one}      & Zero-1-to-3~\cite{liu2023zero} & Objv-40k* & 2$\times$ A10 for 6 days & $\approx$45s  \\

     Ours & VGG-16~\cite{SimonyanZ14a} & Objv-40k & 1$\times$ A40 for 3 days &  $<$10s(R)/$<$20s(G) \\
     \bottomrule
\end{tabular}}
     \vspace{-0.2cm}
     \caption{Training cost and inference speed of single-view-reconstruction methods. ``SPN-Chair'' denotes ShapeNet Chair dataset. ``opt.'' denotes optimization. 
     ``Objv-40k'' denotes a subset from Objaverse with around 40k 3D models.
     `R' and and `G' denote regression-based and generation-based slicing, respectively. Note that the first stage of One-2-3-45 (i.e., Zero-1-to-3) is trained with nearly the whole Objaverse dataset. In the second stage, it is trained with a subset in the scale of 40k 3D models. The inference speed is tested on a single Nvidia-A40-GPU for all methods.} 
    \label{tab:comp_train_cost_infer_speed}
\end{table}

%% file: figs/fig_supp_slicing_canonical_vs_camera.tex
\begin{figure*}[!b]
    \centering
    \includegraphics[width=0.98\textwidth]{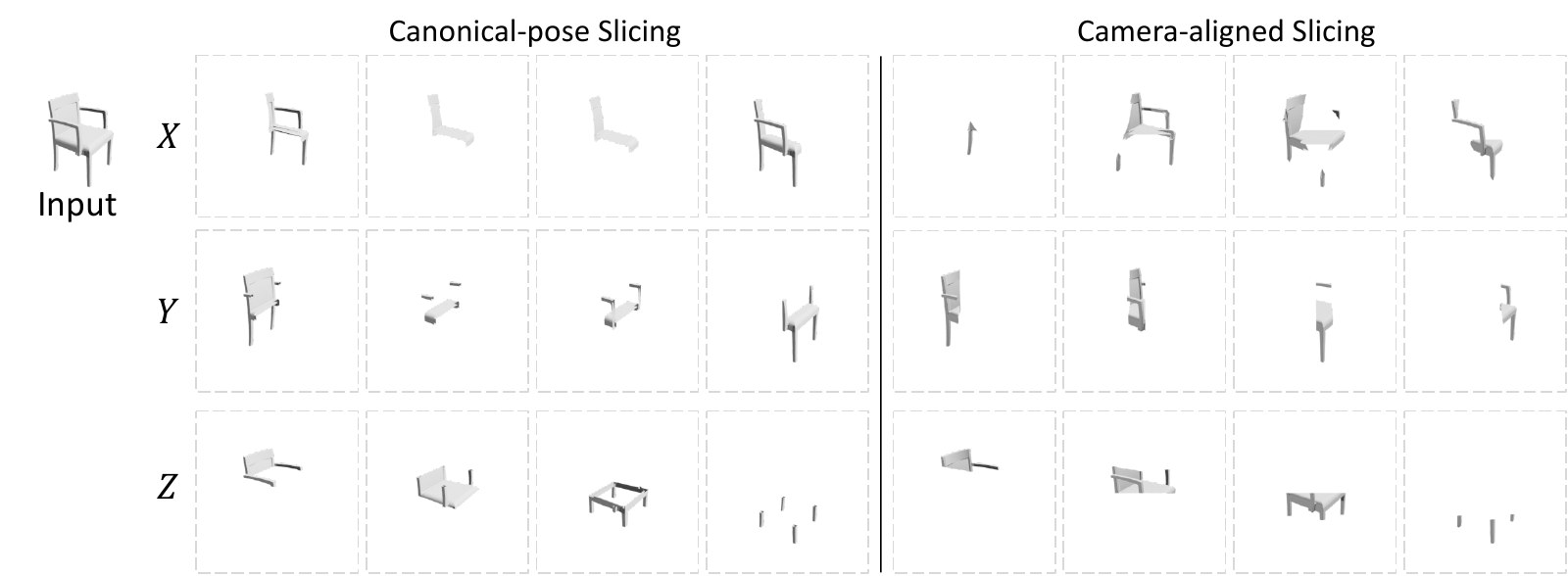}
\caption{Canonical-pose slicing v.s. camera-aligned slicing. In canonical-pose slicing, the slicing directions are determined by the canonical pose of the object. In camera-aligned slicing, the slicing directions are determined by the orientation of the camera.
}
\label{fig:supp_slicing_canonical_vs_camera}
\end{figure*}

%% file: figs/fig_supp_comp_qual_sp_chairs_syncdreamer.tex
\begin{figure*}[!b]
    \centering
    \includegraphics[width=0.98\textwidth]{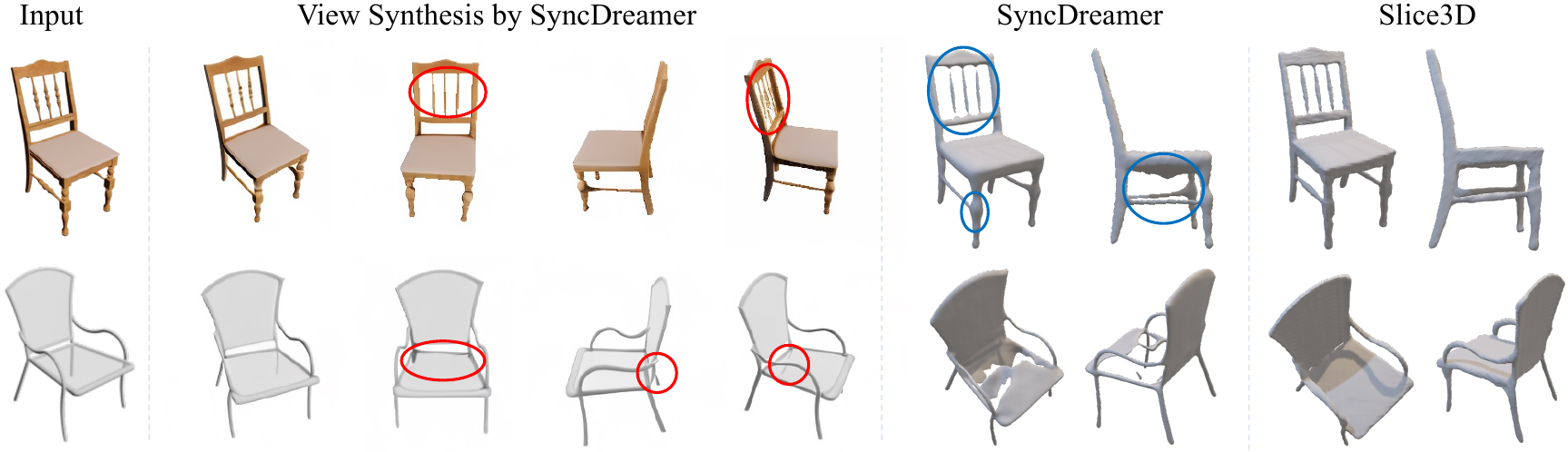}
\caption{Visual comparison against SyncDreamer~\cite{liu2023syncdreamer}, which aims to enhance the consistency by performing spatial attention across different views. The red circles highlight the inconsistency across different synthesized views. For example, the pillars in the first chair and the rear legs in the second chair. The blue circles highlight the artifacts in the 3D mesh.}
\label{fig:supp_comp_qual_sp_chairs_syncdreamer}
\end{figure*}

%% file: figs/fig_supp_slice_visualization.tex
\begin{figure*}[!t]
    \centering
    \includegraphics[width=0.98\textwidth]{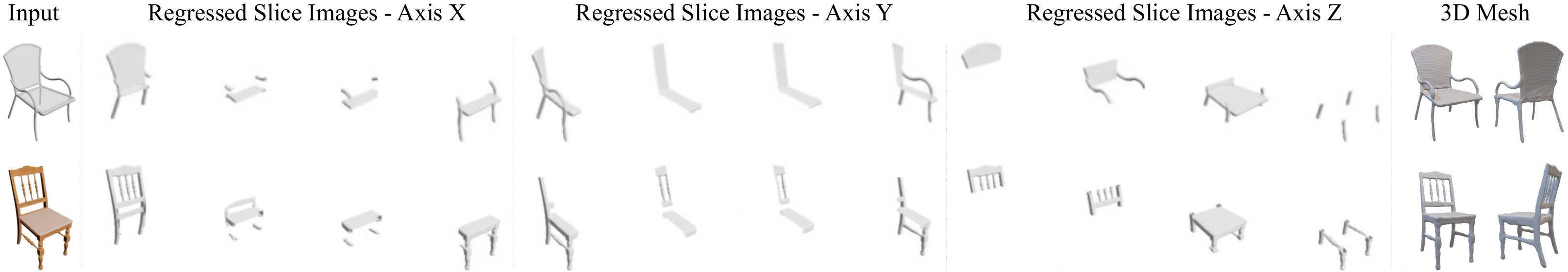}
\caption{Visualization of the predicted (regressed) slice images from the input views.
}

\label{fig:supp_generative_slices_visualization}
\end{figure*}

%% file: tables/tab_supp_slices_concatenation_in_diffusion.tex
\begin{table}[t]
\centering
    \resizebox{0.5\columnwidth}{!}{
    \begin{tabular}{cccc}
     \toprule
     Method               & CD$\downarrow$  & F1$\uparrow$ & HD$\downarrow$   \\ \midrule
     \addlinespace[4pt] 
     Ours (G-C) & 25.0 & 1.51 & 16.4 \\
    Ours (G-S) & \textbf{20.0} & 1.51 &  \textbf{14.1}  \\
    \bottomrule
\end{tabular}}

     \caption{Quantitative results of single-view 3D reconstruction on the Objaverse dataset. `G-C' and `G-S' denote concatenating the slice images along the color and a spatial dimension, respectively.}
    
    \label{tab:slices_concatenation_in_diffusion}
\end{table}

%% file: figs/fig_supp_gen_slices_show_full.tex
\begin{figure*}[!t]
    \centering
    \includegraphics[width=0.98\textwidth]{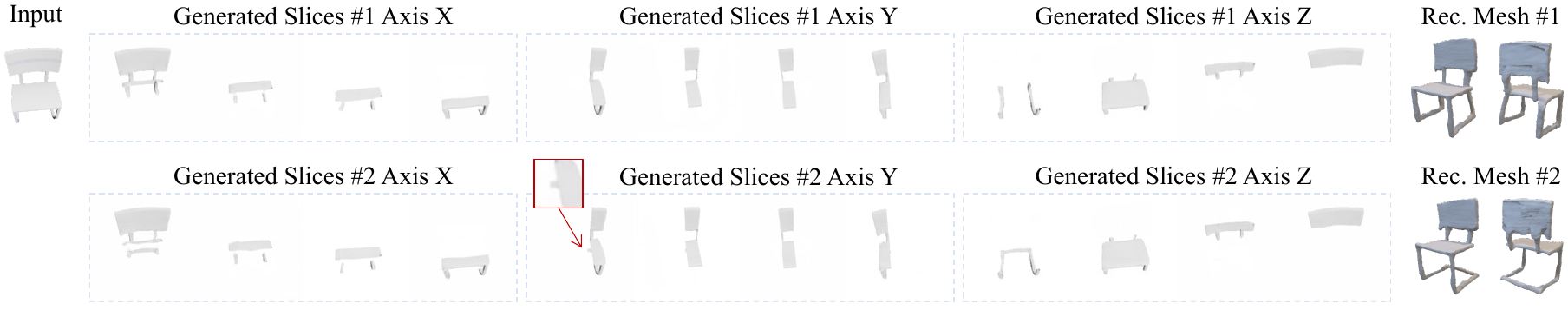}
\caption{Multi-slice generation results and the meshes resulted from them.
}

\label{fig:supp_generative_slices_show_full}
\end{figure*}

%% file: figs/fig_supp_comp_qual_sp_chairs.tex
\begin{figure*}[h]
    \centering
    \includegraphics[width=0.98\textwidth]{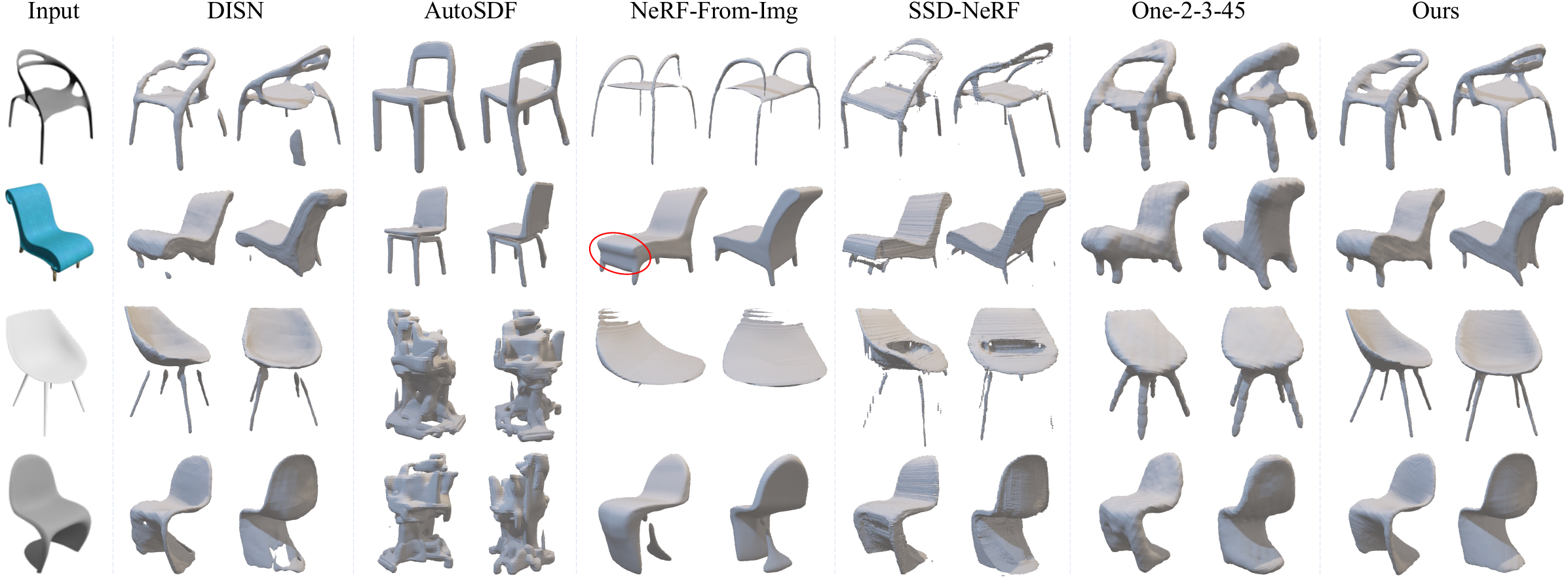}
\caption{More visual comparison between single-view 3D reconstruction methods on ShapeNet chairs. DISN and our method (based on regressive slicing) utilize the same estimated camera parameters. Two different views are displayed to remove view bias.}
\label{fig:supp_comp_qual_sp_chairs}
\end{figure*}

%% file: figs/fig_supp_comp_qual_sp_cars.tex
\begin{figure*}[h]
    \centering
    \includegraphics[width=0.98\textwidth]{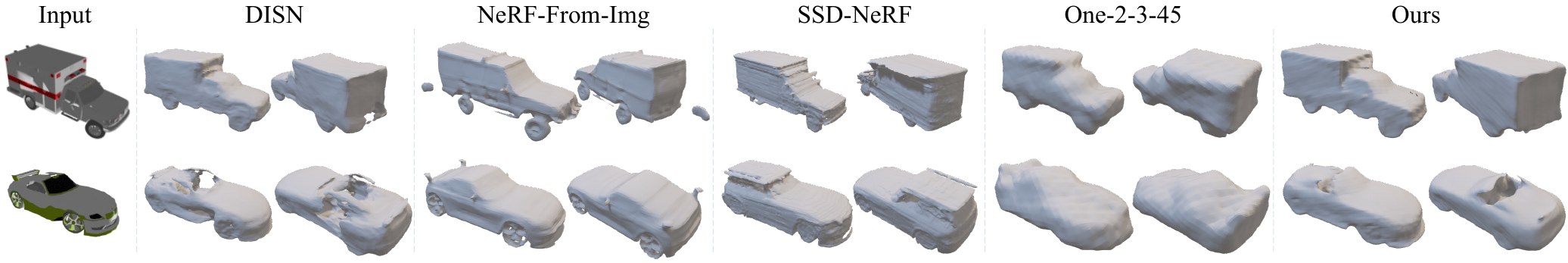}
\caption{Visual comparison between single-view 3D reconstruction methods on two ShapeNet cars.}
\label{fig:supp_comp_qual_sp_cars}
\end{figure*}

%% file: figs/fig_supp_comp_qual_objaverse.tex
\begin{figure*}[h]
    \centering
    \includegraphics[width=0.99\textwidth]{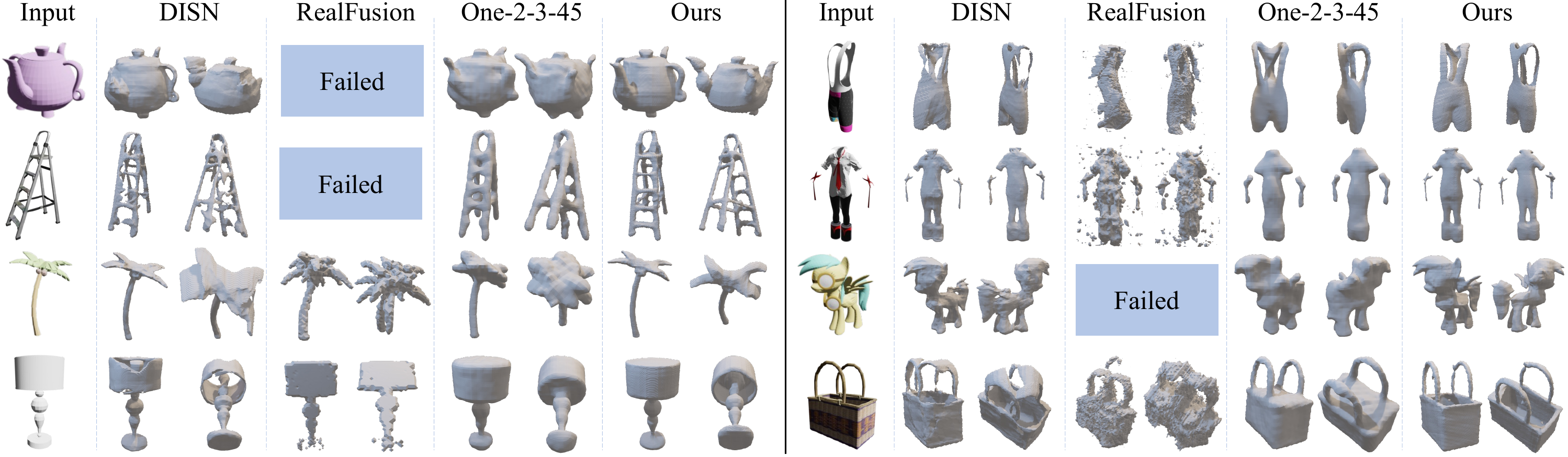}
\caption{More results on Objaverse. ``Failed'' denotes no meaningful results after several optimizations of RealFusion.}
\label{fig:supp_comp_qual_objaverse}
\end{figure*}

%% file: figs/fig_supp_comp_qual_gso.tex
\begin{figure*}[h]
    \centering
    \includegraphics[width=0.99\textwidth]{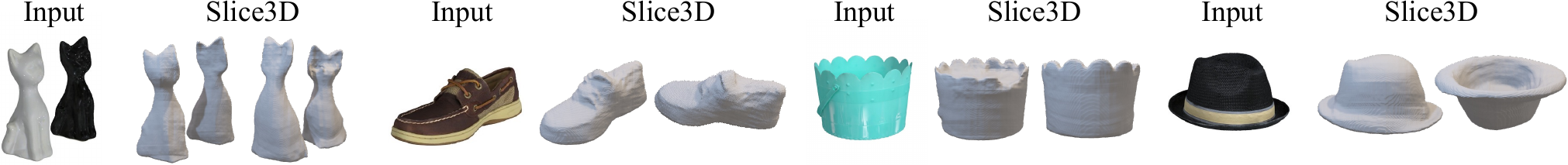}
\caption{More results on GSO ~\cite{downs2022google} dataset.}
\label{fig:supp_comp_qual_gso}
\end{figure*}